\documentclass{article}

\PassOptionsToPackage{numbers,compress}{natbib}

\usepackage[final]{nips_2017}
\usepackage{placeins}
\usepackage{subcaption}
\usepackage{scrextend}

\usepackage[utf8]{inputenc} 
\usepackage[T1]{fontenc}    
\usepackage{hyperref}       
\usepackage{url}            
\usepackage{booktabs}       
\usepackage{amsfonts}       
\usepackage{nicefrac}       
\usepackage{microtype}      

\usepackage[toc,page]{appendix}
\usepackage{bm}
\usepackage{amsmath}
\usepackage{amsfonts}
\usepackage{amssymb}
\usepackage{amsthm} 
\usepackage{mathtools}
\usepackage{booktabs}
\usepackage{makeidx}
\usepackage{sectsty}

\newtheorem*{theorem*}{Theorem}%

\newcommand\Bm{\bm{m}}

\newcommand\BC{\bm{C}}

\newcommand\TR{\mathbf{\mathrm{Tr}}}

\allowdisplaybreaks

\makeindex

\title{Fr\'{e}chet ChemNet Distance: A metric for generative models for molecules in drug discovery}

\author{
 Kristina Preuer \And Philipp Renz \And Thomas Unterthiner \And Sepp Hochreiter \And G\"{u}nter Klambauer \And \\
 LIT AI Lab \& Institute of Bioinformatics, \\
 Johannes Kepler University Linz\\
 A-4040 Linz, Austria\\
\texttt{\{preuer, unterthiner, hochreit, klambauer\}@bioinf.jku.at}
}

\begin{document}
\maketitle

\begin{abstract}
The new wave of successful generative models in machine learning has increased the interest in deep learning driven de novo drug design.
However, assessing the performance of such generative models is notoriously difficult.
Metrics that are typically used to assess the performance of such generative models are
the percentage of chemically valid molecules or the similarity to real molecules in terms of particular
descriptors, such as the partition coefficient (logP) or druglikeness.
However, method comparison is difficult because of the inconsistent use of evaluation metrics,
the necessity for multiple metrics,  and the fact that some of these measures can easily be tricked by simple rule-based systems.
We propose a novel distance measure between two sets of molecules, called Fr\'{e}chet ChemNet distance (FCD),
that can be used as an evaluation metric for generative models.
The FCD is similar to a recently established performance metric for comparing image generation methods, the Fr\'{e}chet Inception Distance (FID).
Whereas the FID uses one of the hidden layers of InceptionNet, the FCD utilizes the penultimate layer of
a deep neural network called ``ChemNet'', which was trained to predict drug activities.
Thus, the FCD metric takes into account
chemically and biologically relevant information about molecules, and
also measures the diversity of the set via the distribution of generated molecules.
The FCD's advantage over previous metrics is that it can
detect if generated molecules are a)~diverse and have similar
b)~chemical and c)~biological properties as real molecules.
We further provide an easy-to-use implementation that only requires the SMILES representation of the generated molecules as input to calculate the FCD.
Implementations are available at: \href{https://www.github.com/bioinf-jku/FCD}{github.com/bioinf-jku/FCD}.
\end{abstract}

\section*{Introduction}
Deep Learning has recently enabled generative models for molecules \citep{segler2017generating}. SMILES strings \citep{weininger1988}
offer a convenient representation of molecules for generative methods. Therefore, Recurrent Neural Networks (RNNs) are a natural basis
for these methods, as they excel at generating text.
RNNs have been combined with variational autoencoders (VAE) \citep{gomez2016automatic}, transfer learning \citep{segler2017generating},
reinforcement learning (RL) \citep{Jaques2016, olivecrona2017molecular} and generative adversarial networks (GAN) \citep{Guimaraes2017} to yield generative models for molecules. Recently, also graph generating methods
have been suggested \citep{li2018learning,simonovsky2018graphvae} which can be used to generate graph representations
of molecules.

Up to now, there is no established evaluation criterion that is consistently used across all publications to assess generative models for molecules.
The most basic and commonly reported metric is the fraction of valid SMILES, which is a prerequisite for a generative model.
However, this metric can easily be maximized by generating simple molecules, such as ``CC'' and ``CCC'', with a rule-based system.
Other metrics focus on numerical or visual comparison of molecular properties like solubility \citep{olivecrona2017molecular},
number of rotational bonds \citep{olivecrona2017molecular}, number of aromatic rings \citep{olivecrona2017molecular}, synthetic
accessibility \citep{Guimaraes2017, Jaques2016, Tsuda2017} and druglikeness \citep{Guimaraes2017, Jaques2016}. Additional metrics that are used aim at assessing the diversity of the generated molecules such as calculating the Tanimoto similarity/distance
to the training set \citep{Guimaraes2017, segler2017generating} or within the generated molecules \citep{benhenda2017chemgan}. The
diversity of recently developed methods for molecule generation shows the high interest in this new research field, but the diversity
of different evaluation methods inhibits focused and comparative research. Therefore, it is necessary to find a common basis for evaluating generative models for molecules. We will suggest such a metric and show that previously used metrics have their specific flaws, and fail to detect certain biases in generative models.

A proficient metric should capture the validity, chemical and biological meaningfulness, and diversity of the generated molecules.
We therefore aim at creating a metric which
is capable of unifying these requirements into one score. To this end we adopt a strategy that
has been established to compare generative models for images, the Fr\'{e}chet Inception Distance (FID) \citep{heusel2017gans}.
This distance measure uses a representation obtained from the Inception network\citep{salimans2016improved} to represent the input objects
of the network. For these image representations, the Fr\'{e}chet distance is used as a distance measure.
Analogously, we use the hidden representation from a
neural network, called “ChemNet” (see details below), which was trained to predict biological
activities, as representation of molecules. This representation contains both chemical as well
as biological information about the molecules, since the input layer can be seen as purely
chemical information and the output layer as purely biological information. The layers in
between contain both chemical and biological information about the molecule, which is a
representation that we desire. By taking into account the distribution of these representations
within a set of molecules, we also capture the diversity within the set. 
A purely chemical representation could be also used in combination with the Fréchet distance. This representation could be a set of fingerprints, the principle components calculated on the fingerprints, the latent space of an autoencoder to name just a few. However, we hypothesize that using just the chemical information is not sufficient to judge generative models for drug design. We will examine our assumption by comparing the FCD to a fingerprint based Fréchet distance, which we call Fréchet Fingerprint Distance (FFD).

\paragraph{Fr\'{e}chet ChemNet Distance.}
We introduce the  Fr\'{e}chet ChemNet Distance (FCD) to calculate the distance between the distribution $p_w(.)$ of 
real-world molecules and the distribution $p(.)$ of molecules from a generative model. To obtain a numerical representation 
of each molecule, we use the activations of the penultimate layer of ``ChemNet'' (see next paragraph). We then calculate the 
first two moments (mean, covariance) of these activations for each of the two distributions. Since the Gaussian is the maximum
entropy distribution for given mean and covariance, we assume the hidden representations to follow a multidimensional Gaussian.
The two distributions ($p(.)$, $p_w(.)$) are compared using the Fr\'{e}chet distance \cite{Frechet:57}, which is also 
known as Wasserstein-2 distance \cite{Wasserstein:69}. We call the Fr\'{e}chet distance $d(.,.)$ between the Gaussian $p_w(.)$ with mean and covariance
$(\Bm_w,\BC_w)$ obtained from the real-world samples and the Gaussian  $p(.)$ with mean and covariance $(\Bm,\BC)$ 
obtained from a generative model  the ``Fr\'{e}chet ChemNet Distance'' (FCD), which is given by \cite{Dowson:82}:
\begin{align}
 d^2((\Bm,\BC),(\Bm_w,\BC_w))=\|\Bm-\Bm_w\|_2^2+  \TR \bigl(\BC+\BC_w-2\bigl(
\BC\BC_w\bigr)^{1/2}\bigr).
\end{align}
Throughout this paper, the FCD is reported as $d^2(.,.)$ analogously to \citet{heusel2017gans}.

The FCD is based on the activations of the penultimate layer of ChemNet. The model \citep{Mayr2018} was 
trained to predict bioactivities of about 6\,000 assays available in three major drug discovery databases (ChEMBL \citep{Bento2013}, ZINC \citep{Irwin2012}, PubChem \citep{Wang2016}).
We used long short-term memory (LSTM) \citep{Hochreiter1997} recurrent neural networks based
on the one-hot encoded SMILES representation of chemical molecules.
The full architecture of this model consists of two 1D-convolutional layers with SELU
activations, \citep{Klambauer2017} followed
by a max-pooling layer, two stacked LSTM layers,
and a fully connected output layer.
Hyperparameter selection and training were done on two thirds of the available data and the
last third was used for testing. The network was optimized for mean predictive performance. The hidden representation of the 2nd LSTM layer after processing the full input sequence is used for calculating the FCD.

To estimate the mean $\Bm$ and covariance $\BC$ of the real and generated samples sufficiently
large data sets should be chosen. Therefore, we determined the necessary sample size for a reliable estimate. We have chosen 200\,000 randomly drawn real molecules to represent the reference molecules. Additionally, we have drawn 5, 50, 500, 5\,000, 50\,000 and 300\,000 real molecules which represent different data sizes of
generated samples. If the data set size of the ``generated'' samples is sufficiently large than the
mean $\Bm$ and covariance $\BC$ can be accurately estimated and the FCD is close to 0.
This experiment was repeated 5 times, to show that for a large enough sample the variance of
FCD becomes negligible. For 5, 50, 500, 5\,000, 50\,000 and 300\,000 samples the mean FCD values~$\pm$~ one standard deviation were 76.46$\pm$5.03, 31.86$\pm$0.75, 4.41$\pm$0.03, 0.42$\pm$0.01, 0.05$\pm$0.00, 0.02$\pm$0.00, respectively. Therefore, a sample size of 5\,000 is already sufficient to get a mean FCD which is close to 0 and a negligible variance meaning that the
mean $\Bm$ and covariance $\BC$ can be accurately estimated.

\paragraph{Detecting flaws in generative models.}
We determine the capability of the FCD to detect if a generative model has produced diverse
molecules which possess chemical and biological properties similar to already
known molecules. We compare the FCD to four commonly reported metrics: mean logP \citep{Wildman1999},
mean druglikeness \citep{Bickerton2012}, mean Synthetic Accessibility (SA) score \citep{Ertl2009} and the internal diversity score with Tanimoto
distance \citep{benhenda2017chemgan} of the generated samples. Additionally, to evaluate if the activations of ChemNet are a suitable representation for molecules we also compare the FCD to the Fr\'{e}chet Fingerprint Distance (FFD) which is calculated analogously to the FCD. However, in contrast to the FCD, the FFD is based on 2048 bit ECFP\_4 fingerprints. A generative model performs well if it produces samples in a similar logP and druglikeness range as the training data and possesses comparable
internal diversity.

All compared metrics were calculated in RDKit. The internal diversity score is calculated using 
Morgan Fingerprints of RDKit with radius 2 (equivalent to ECFP\_4 \citep{Rogers2010}).
For the methods comparison we have selected a subset of real molecules from the three databases which were neither used for training ChemNet nor to estimate the real sample statistics ($\Bm$,~$\BC$) needed for calculating the FCD.

For the real subsample of the combined databases we calculated each metric, i.e., logP, druglikeness, SA score, internal diversity\footnote{The calculation of the full distance matrix required for this
metric is too time and memory intense, therefore we have averaged the results from 5 subsets of randomly drawn 5\,000 samples.}, FFD and  FCD to illustrate the baseline value for real samples. 
In the next step, to simulate generative models with particular flaws, 
we created ``disturbed data sets'' of molecules with a) low druglikeness ($<$~5th~percentile), b)~high
logP values ($>$~95th~percentile), c)~low SA scores ($<$~5th~percentile), d)~high Tanimoto similarity and e)~stemming from the same target class to simulate generative models producing molecules a) with a low druglikeness, b)~a high logP, c)~which are difficult to synthetize, d)~having low diversity and e)~are active for a specific target.
We aim to show that FCD can detect all four biases, thereby combining the benefits of other metrics into a single one and is able to identify distribution differences in biological meaningful subsets.

\begin{itemize}
 \item {\bf Bias towards low druglikeness:} A biologically based assessment can be performed in terms of the average druglikeness of the generated compounds. 
Druglikeness \citep{Bickerton2012} is the geometric mean of several desired molecular properties
such as solubility, permeability and metabolic stability. Therefore, it is often used to
determine if generated molecules are close to real samples. Molecules with a low druglikeness
are commonly not desired since they possess low bioavailability. We simulated models generating molecules with a druglikeness lower than the 5th percentile of the druglikeness values
(i.e.~$0.19$) of real molecules. We randomly selected 5\,000 molecules with a low druglikeness for this simulation.
This was repeated 5 times to create 5 different generative models producing molecules with a low druglikeness.

 \item {\bf Bias towards high logP:} Similar chemical properties such as the average logP value of the generated molecules are another 
 way to judge whether created molecules are reasonable or not.
Therefore, we simulated a generative model that has a
bias towards generating molecules with a high logP. For this purpose, we selected 5\,000 molecules that
have a logP value higher than the 95th percentile of logP values (i.e.~>~$6.15$) of real molecules. Although these
molecules are valid, they are located on the edge of the logP distribution, which
should be detectable by an appropriate metric. This procedure was repeated 5 times to simulate
5 different generative models with a bias towards molecules with a high logP.

 \item {\bf Low synthetic accessibility:} Furthermore, it is beneficial if a generative model produces molecules which are indeed synthetically feasible. Therefore, we simulated generative models which have a bias towards molecules with a low synthetic accessibility (SA) score. We randomly drew 5\,000 molecules with a SA score lower than the 5th percentile of the SA scores (i.e.~<$1.91$) of real molecules. This procedure was repeated 5 times to simulate
5 different generative models with a bias towards low SA.

 \item {\bf Mode collapse:} Another desirable property of generative models is to create a wide variety of different samples.
However, generative models might suffer from mode collapse \citep{Metz2017, unterthiner2017coulomb}, such that they produce
only molecules with a low diversity. Although there also exist models which do not suffer from mode collapse \citep{Popova2017}, an appropriate metric should be able to assess
the internal diversity of the generated molecules. We simulated generative models which suffer from mode collapse. For this
purpose, we used single linkage clustering with a Tanimoto similarity cut-off of 0.65. We have chosen a large cluster of which we randomly selected 5\,000 molecules. This procedure was repeated 5 times to simulate 5 different generative models suffering from mode collapse.

 \item {\bf Kinase inhibitors:} Conditional generative models are used to produce molecules active for a specific target. We assessed if the metrics can catch the bias towards a certain target family. For this experiment we have selected a large scale activity assay from PubChem (AID 720504). We have randomly selected 5\,000 active molecules for polo-like kinase 1 - polo-box domain (PLK1-PBD).

\end{itemize}

\begin{center}
\begin{figure}
\includegraphics[width=\textwidth]{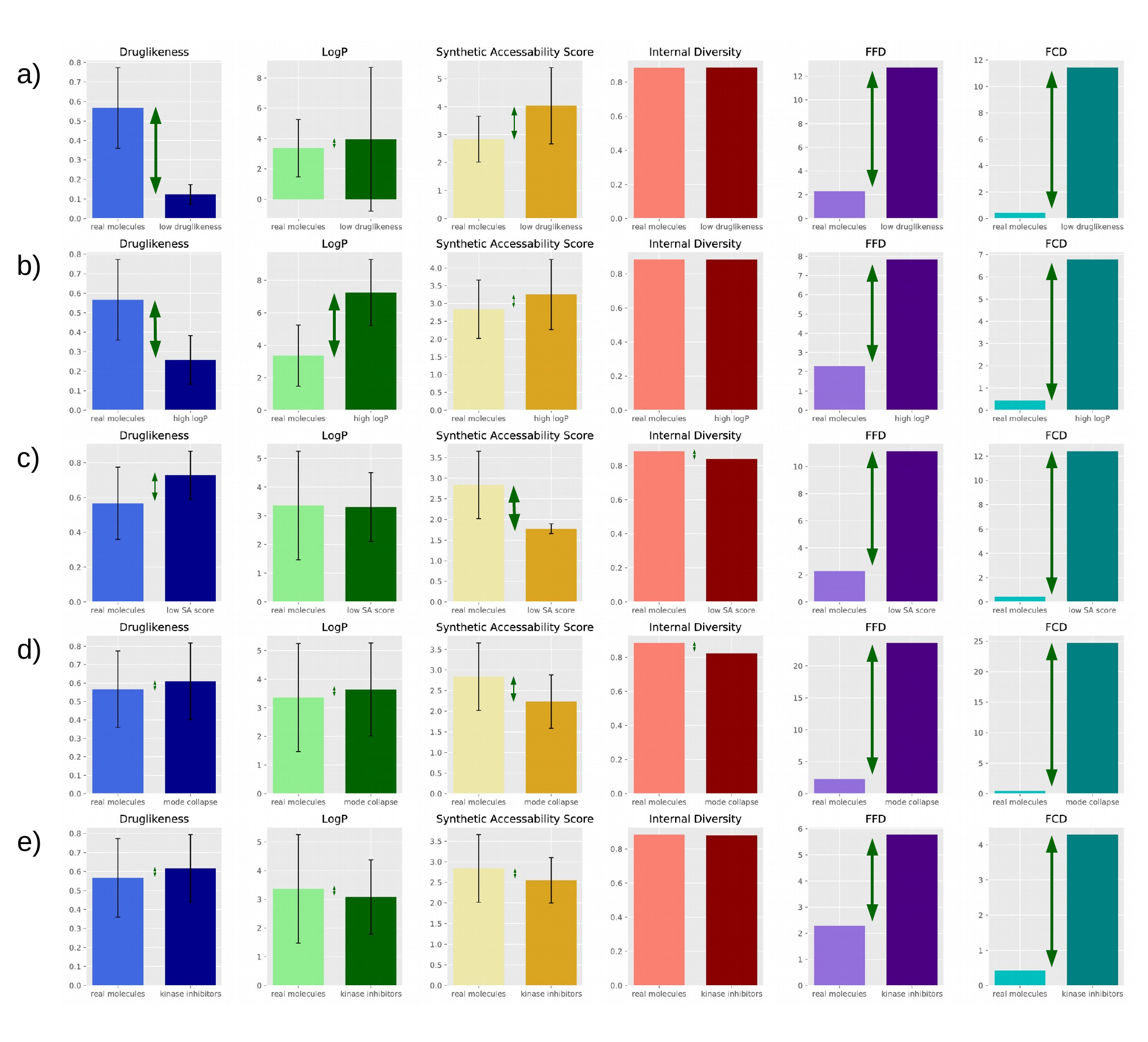}
\caption{\label{fig:comparison} Mean responses of  druglikeness (blue), logP (green), synthetic accessibility score (yellow), internal diversity (red), FFD (purple) and
FCD (cyan) to the real sample data set and the disturbed molecule sets a)~with  low druglikeness, b)~a high logP, c)~which are difficult to synthetize, d)~having low diversity and e)~stemming from one target family. The height of the bars indiciate the mean of the metrics over repeated experiments and the error bars indicate the mean standard deviation across molecules of the repeated experiments. Please note that for the internal diversity, the FFD and the FCD the standard 
deviation cannot be determined because these measures are based on sets and cannot be determined for single molecules. 
The green arrows indicate the difference between the mean measure of the real set and the disturbed set. In a),~the low druglike set is clearly detected by druglikness, the FFD and the FCD. In b),~logP, druglikeness, the FFD and the FCD detect the bias towards high logP
molecules. In c),~the bias towards molecules with a low SA score is clearly revealed by the SA score, the FFD and the FCD.
In d),~the models suffering from mode collapse are discovered by the internal diversity, the FFD and the FCD. In e) the active molecules for PLK1 are uncovered by the FFD and the FCD.}
\end{figure}
\end{center}

We examined the average changes of the metrics when generative models suffer from
a)~a bias towards low druglikeness, b)~a bias towards high logP molecules, c)~a bias towards low SA scores and d)~mode collapse. Furthermore we assess the changes of the metrics for e) molecules stemming from the same target class (see Fig.\ref{fig:comparison}). In a),~druglikeness, the FFD and the FCD are able clearly to detect the disturbed data set. In b),~the bias towards high logP values is reflected in the druglikeness, LogP, the FFD and the FCD. In c),~ the disturbed set is recognized by the SA score, the FFD and the FCD and to a minor extend by druglikeness and internal diversity. The mode collapse of d) is clearly revealed by internal diversity, the FFD and the FCD. The active compounds for PLK1 are only uncovered by the FFD and the FCD. Both FFD and FCD show in all four cases a clear difference in the mean of the disturbed and undisturbed data sets. The difference is especially high for both metrics in the case in which we simulated a generative model that suffers from mode collape. Although, FFD is able to detect all the biased sets, FCD is able to make more distinct differentiations. Especially in epxeriment e) where more biological relevant information is necessary, the FCD shows superior behaviour.

\FloatBarrier

\paragraph{FCD of recent generative models.}
We calculated the FCD for publicly available SMILES strings of recently developed generative
models \citep{benhenda2017chemgan, olivecrona2017molecular, segler2017generating}. \citet{segler2017generating} generated more than 450\,000~SMILES
strings with an LSTM network used for next character prediction. \citet{benhenda2017chemgan}
aimed at producing SMILES strings with ORGAN \citep{Guimaraes2017} and RL that are active
against the dopamine receptor D2 (DRD2). 32\,000 molecules were generated after 40 and 60
training iterations for RL, and after 30 and 60 training iterations for ORGAN. \citet{olivecrona2017molecular} trained two RL
agents to produce molecules active for the DRD2 receptor. The canonical and
the reduced agent were trained on the complete ChEMBL and a subset from which molecules similar to
Celecoxib were removed, respectively. Each agent produced 128\,001 molecules.
Furthermore, we examined a simple rule-based approach, which randomly draws C, N and O atoms and concatenates them to obtain SMILES with random lengths between 1 and 50. 
The sample statistic was calculated with 200\,000 randomly selected real molecules that were not used for training ChemNet. For each model, we have randomly drawn 10\,000 samples of the generated SMILES 10 times and determined the FCD score.

\begin{center}
\begin{figure}

	\includegraphics[width=\textwidth]{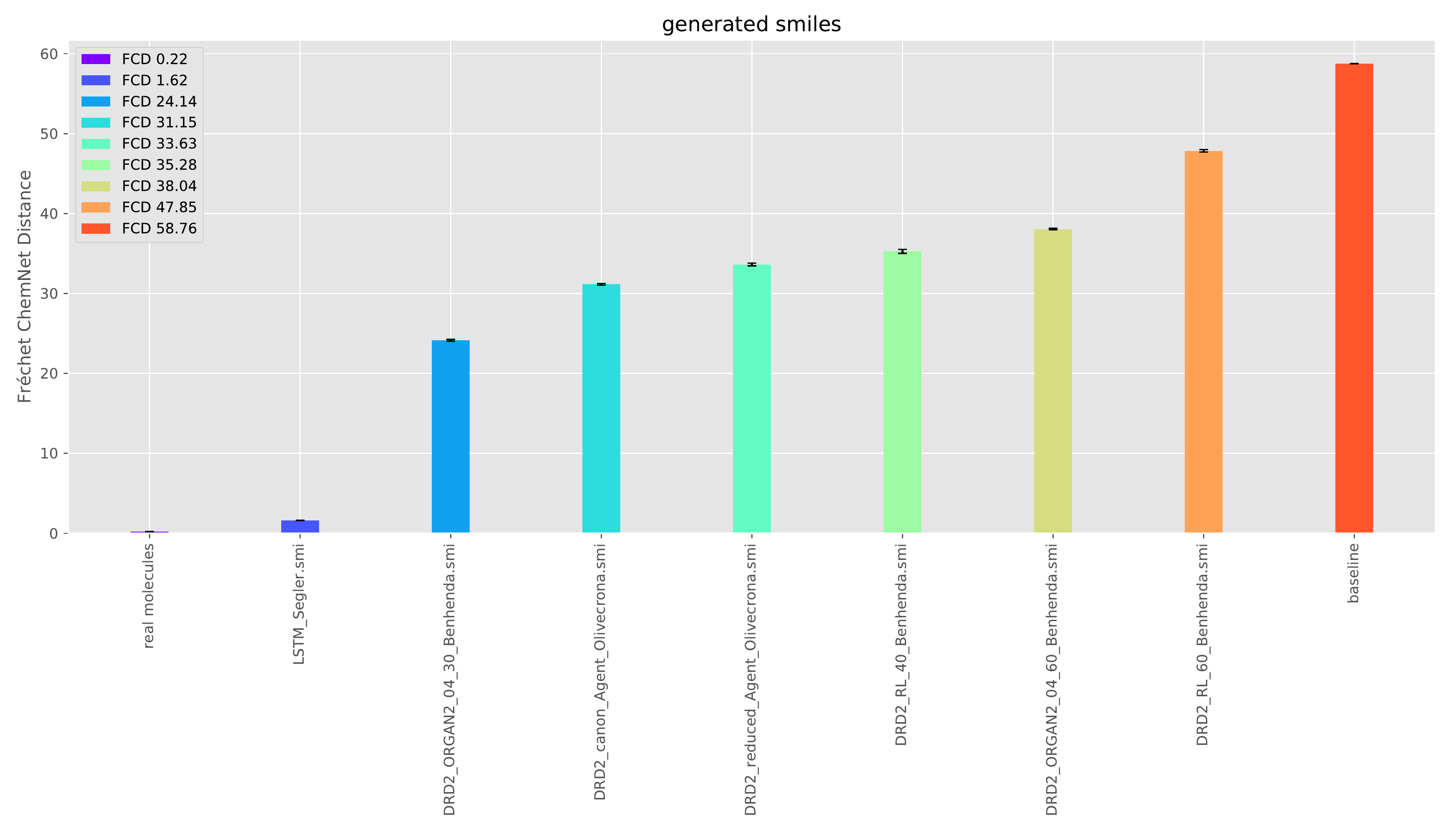}
	\caption{FCD of generated molecules based on different methods. A low value indicates that the distribution of the real molecules
	was approximated well by the generated molecules. The method ``Segler'' was optimized to produce molecules similar to the molecules
	present in the ChEMBL database. Six further models were optimized to produce molecules with activity on
	a certain protein target. This optimization process leads to a higher FCD value indicating a larger difference
	to the distribution of real molecules.\label{fig:methods}}
\end{figure}
\end{center}

We calculated our new FCD performance metric for the following methods:
\begin{itemize}
 \item {\bf Segler:} The next-character LSTM approach by \citet{segler2017generating}.
 \item {\bf Olivecrona canon agent:} An RNN-based RL approach to generate active molecules for DRD2 \citep{olivecrona2017molecular}.
 \item {\bf Olivecrona reduced agent:} An RNN-based RL approach to generate active molecules for DRD2 using a reduced training set \citep{olivecrona2017molecular}.
 \item {\bf RL 40/60 iterations:} An LSTM-based RL approach to generate active molecules for DRD2 trained for 40/60 iterations \citep{benhenda2017chemgan}.
 \item {\bf ORGAN 30/60 iterations:} A GAN using an LSTM-based generator and a CNN-based discriminator trained for 30/60 iterations \citep{benhenda2017chemgan}.
  \item {\bf baseline:} A generative model producing just methane.
  
\end{itemize}
Fig. \ref{fig:methods} shows the results for the different methods. Please note that this is not a methods comparison but should demonstrate that the ranking 
by the FCD matches our intuition. The lowest FCD value of $0.22$ is obtained by randomly drawn real molecules. Since this value is close to zero the
randomly drawn subsamples sufficiently represent the underlying distribution of real molecules. The method ``Segler'' \citep{segler2017generating} achieved 
the second lowest FCD of $1.62$, indicating that the distribution of these generated molecules is closer to the real molecule distribution than distributions produced by other methods. This matches our intuition, because the other methods (``Olivecrona'', ``RL'', and ``ORGAN'') were optimized to generate
molecules that are active for DRD2 and are
therefore not designed to approximate the distribution of the complete set of real molecules. This optimization procedure is clearly 
captured by the FCD metric: for all these methods the FCD is notably higher, ranging from $24.14$ to $47.85$.
Furthermore, ORGAN after 60 iterations and  RL after 60 iterations produce molecules that are more distant from real molecules 
than ORGAN after 30 iterations and  RL after 40, respectively. Intuitively, more training iterations lead to more DRD2 specific molecules and 
therefore to molecules more distant to the complete real molecule distribution and lower diversity \citep{benhenda2017chemgan}. Additionally, the
FCD captures that the canonical and the reduced agents both learn a similar chemical space as concluded by \citet{olivecrona2017molecular}. 
The rule-based system has the highest FCD of $58.76$, which can be considered an easily achievable  baseline.
Overall, the ranking of the methods by their FCD matches our intuition and previous findings.
In this comparison, randomly drawn molecules from the combined data set were used to determine the sample statistic underlying FCD, therefore methods which are
optimized to capture the distribution of active molecules for DRD2 have a higher FCD. By using a
different distribution to calculate the sample statistic, the FCD could also be used to evaluate targeted molecule generation.

\paragraph{Conclusions.}
In previous studies, the assessment of generative molecules was based on specific properties such as logP, druglikeness or SA score. However, looking at all these properties individiually makes the comparison of generative models difficult. We introduce the FCD, a novel metric for generative models for drug design. FCD is based on a multi-task network and therefore incorporates a wide variety of important chemical and biological features into a single metric. 
FCD was able to detect four potential flaws of generative models, as we have demonstrated in our experiments. Furthermore we also show that the FCD can also catch a biological bias (active PLK1 kinase inhibitors). 
Our proposed approach is not restricted to generative models that produce SMILES strings, but can readily be used
for graph generating methods by converting the produced molecules into a SMILES format. Within our experiments, we compare the FCD also to a fingerprint based Fréchet distance. This comparison clearly illustrates that by incorporating biological information the metric further improves and the differences between the real and biased sets are more distinct. Overall, we show that FCD is a comprehensive, simple and powerful metric for the evaluation of generative models in drug discovery. 

\section*{Availability}
Implementations to calculate FCD and to reproduce the experiments of this work
are available at: \href{https://www.github.com/bioinf-jku/FCD}{github.com/bioinf-jku/FCD}

\section*{Acknowledgments}
We thank Marwin Segler, Marcus Olivecrona and Mostapha Benhenda for providing their generated molecules. Further more we thank Marwin Segler for helpful discussion. KP, TU, and GK funded by the Institute of
Bioinformatics, Johannes Kepler University Linz Austria.

This work was supported by Merck Group (research agreement 05/2016), Zalando (research agreement 01/2016) and by LIT (LIT-2017-3-YOU-003).

\section*{References}

\bibliography{fcd}
\bibliographystyle{apalike}

\end{document}